\documentclass[runningheads]{llncs}
\usepackage[T1]{fontenc}
\usepackage{graphicx}
\usepackage{amsmath}
\usepackage{times}
\usepackage{booktabs} 
\usepackage{multirow}
\usepackage{enumitem}
\usepackage{xcolor}
\usepackage{hyperref}
\hypersetup{
  colorlinks,
  citecolor=blue,
  linkcolor=red,
  urlcolor=blue}
\usepackage{wrapfig}
\usepackage{marvosym}

\begin{document}
%
\title{NODER: Image Sequence Regression Based on \\Neural Ordinary Differential Equations}
\titlerunning{NODER}
%

\author{Hao Bai \and
 Yi Hong\textsuperscript{\Letter} }   
\authorrunning{Bai and Hong}
%
\institute{
Department of Computer Science and Engineering, \\Shanghai Jiao Tong University, Shanghai, 200240, China \\ \textsuperscript{\Letter} \email{yi.hong@sjtu.edu.cn}}
%
\maketitle              
\begin{abstract}
Regression on medical image sequences can capture temporal image pattern changes and predict images at missing or future time points. However, existing geodesic regression methods limit their regression performance by a strong underlying assumption of linear dynamics, while diffusion-based methods have high computational costs and lack constraints to preserve image topology. In this paper, we propose an optimization-based new framework called NODER, which leverages neural ordinary differential equations to capture complex underlying dynamics and reduces its high computational cost of handling high-dimensional image volumes by introducing the latent space. We compare our NODER with two recent regression methods, and the experimental results on ADNI and ACDC datasets demonstrate that our method achieves the state-of-the-art performance in 3D image regression. Our model needs only a couple of images in a sequence for prediction, which is practical, especially for clinical situations where extremely limited image time series are available for analysis. Our source code is available at \url{https://github.com/ZedKing12138/NODER-pytorch}.


\keywords{Image sequence regression  \and Neural ordinary differential equations  \and Image prediction and generation}
\end{abstract}
\section{Introduction}
In medical image analysis, image sequences like longitudinal image scans or image time series provide rich spatio-temporal information for studying the mechanisms of human aging and the patterns of disease development. Regression on temporal image sequences~\cite{niethammer2011geodesic,hong2014time,ding2019fast,hazra2019spatio} is a commonly-used technique to explore the relationship between images and their associated time attribute. However, in practice, regression on medical image sequences, especially longitudinal 3D image volumes, is facing the following three challenges: (i) {\it Missing data}. Collecting regular follow-up scans of a subject is a challenging task. Often, we have missing scans at one or more time points for each subject. (ii) {\it High-dimension low-sample size data}. In this paper, we tackle 3D medical image sequences, and each volume is high-resolution three-dimensional images with millions of voxels, while each sequence has only tens of image scans for regression. (iii) {\it Semantic richness but with subtle temporal changes}. Each volume has detailed spatial information about tissue structures, which is non-trivial to generate; at the same time, the temporal changes of these tissues are often subtle, which is difficult to capture without a special design or treatment to model the temporal dynamics.   

To address the above challenges, there are two categories of regression approaches, i.e., the optimization-based methods like geodesic regression~\cite{niethammer2011geodesic,ding2019fast,hong2012simple} and the learning-based methods like regression based on diffusion models~\cite{yoon2023sadm,pinaya2022brain}. Geodesic regression extends linear regression to Riemannian manifolds, which is developed in the framework of Large
Deformation Diffeomorphic Metric Mapping (LDDMM)\cite{beg2005computing}. By generalizing diffeomorphic image registration to temporal image data, the regression can compactly model the spatial deformations over time~\cite{beg2005computing,niethammer2011geodesic}.  However, solving the underlying optimization problem is computationally expensive. Therefore, a simplified approximation method, i.e., Simple Geodesic Regression (SGR)~\cite{hong2012simple}, has been proposed, which decouples the iterative optimization of regressing geodesics into pairwise image registrations. To further reduce the computational time with the help of deep learning techniques, the Fast Predictive Simple Geodesic Regression (FPSGR)~\cite{ding2019fast} is proposed by utilizing a fast predictive registration method. Although this method is computationally efficient, its regression accuracy is limited by its assumption of linear temporal changes.

The diffusion-based model is a recent popular alternative to generate high-quality images. One recent image regression method is the Sequentially Aware Diffusion Model (SADM)~\cite{yoon2023sadm}, which augments diffusion models with a sequence-aware transformer as a conditional module. Like geodesic regression, diffusion-based methods can also handle the missing data issue and allow for autoregressive image sequence generation during inference. However, diffusion-based methods likely introduce unwanted structures into the generated images since they are learning-based techniques and have no constraints like diffeomorphic deformations in geodesic regression to ensure the topological preservation and differential homeomorphism properties of the generated images. Other limitations are the requirements of massive data, a long training process, extensive memory usage, and high time consumption during inference.


In addition to the previously mentioned regression methods, several time series modeling approaches utilize spatio-temporal transformers combined with attention mechanisms~\cite{ahn2023star,mazzia2022action}. However, these methods are associated with substantial computational overhead and lack the ability to directly constrain diffeomorphism, making them more suitable for simpler tasks like human action recognition rather than for reconstructing high-resolution 3D medical images. Alternatively, Generative Adversarial Network (GAN) based methods with attribute embedding~\cite{xia2021learning} convert the generation of medical image time series into a multivariable autoregressive problem. Despite this, GAN methods often face convergence challenges during training, and the constraints imposed by reconstruction loss terms to maintain subject identity are limited.

\begin{figure}[t]
\includegraphics[width=\textwidth]{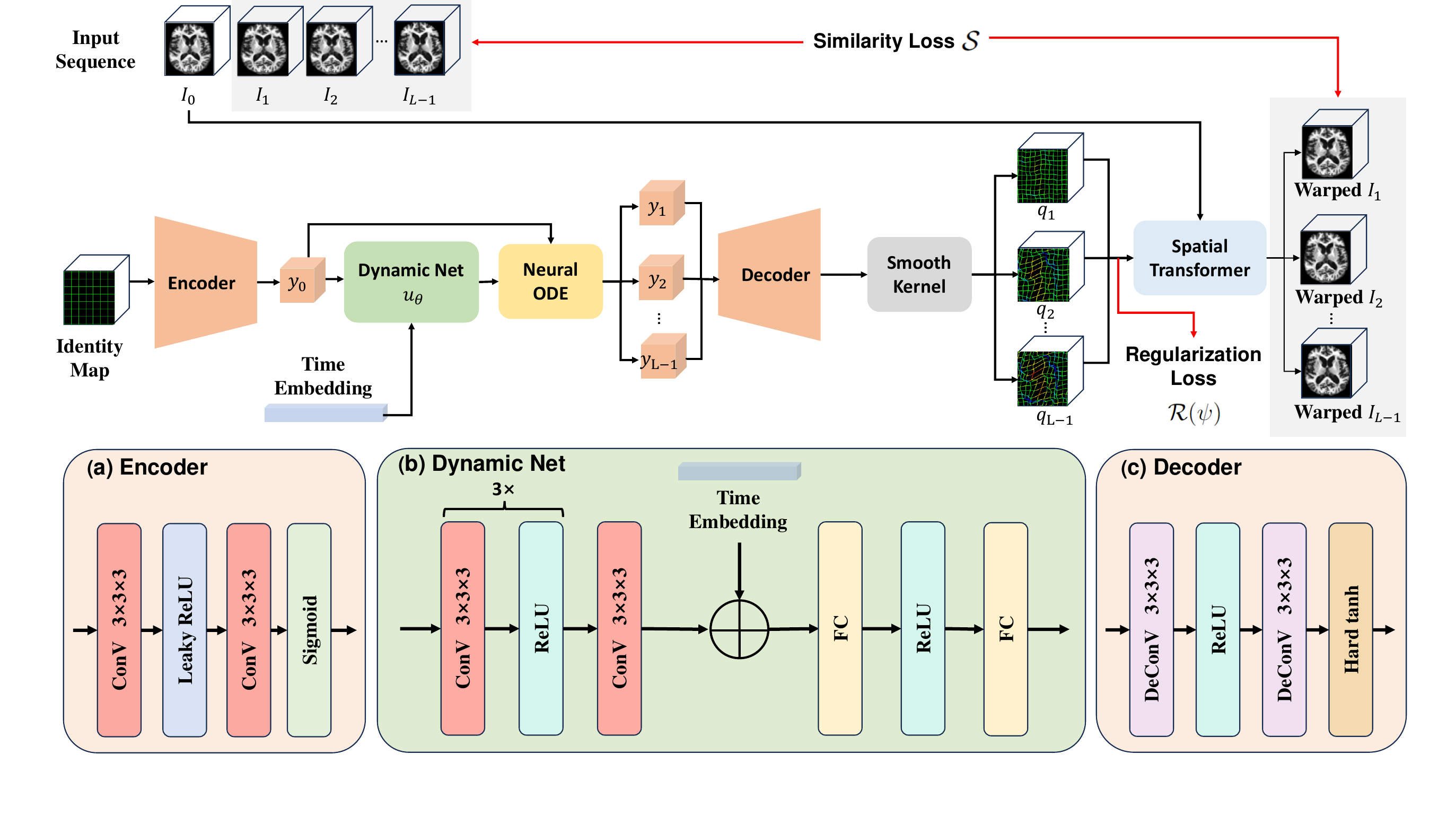}
\caption{Overview of our proposed NODER framework.} \label{fig:overview}
\end{figure}

Therefore, we stick to the optimization-based methods like geodesic regression by using diffeomorphic deformations to drive the image generation over time, but relax its linear dynamic assumption to model more complex dynamics. Fortunately, the neural ordinary differential equations (Neural ODEs)~\cite{chen2018neural} provide a neural network based solution for addressing numerous dynamic fitting problems, which is successfully adopted to model deformable image registration, such as NODEO proposed in~\cite{wu2022nodeo}. Inspired by NODEO, we generalize the Neural ODEs to the image space and handle image regression on a couple of images in a sequence via neural network based optimization. In particular, we propose a model called NODER, which converts the velocity field optimization problem in image regression into a parameterized neural network optimization problem, as shown in Fig.~\ref{fig:overview}. To address the high-computational cost issue faced by Neural ODEs when handling high-dimensional image volumes, we propose to bring the dynamic optimization of Neural ODEs into the latent space via the auto-encoder technique~\cite{shu2018deforming}.
Our contributions in this paper are summarized below:
\begin{itemize}[noitemsep, topsep=0pt]
    \item We propose a novel optimization-based image regression model, NODER, based on Neural ODEs and diffemorphic registration. Our NODER has the freedom to capture complex temporal dynamics in 3D medical image sequence with a couple of images and even missing time points.   
    \item  We conduct experiments on both 3D brain and cardiac MRI datasets. Our NODER generates 3D images with the best quality, compared to recent methods FPSGR and SADM; and it outperforms the diffusion model SADM in terms of the both image quality and the computational cost. 
\end{itemize}

\section{Method}

\subsection{Background and Definitions}
We consider an unparameterized 3D image as a discrete solid, where the position of its $i$-th voxel can be represented as: $x_{i}\in\Omega\subseteq{R}^{3}$,
where $\Omega$ represents the 3D image domain.
The positions of all voxels in the image can be represented by an ordered set:
$q=\{x_i\}_{i=1}^{N}$. Here, $N=D\times H \times W$ represents the total number of voxels in the image, and $D$, $H$, $W$ denote the depth, height, and width of the image, respectively. We denote the domain where the voxel cloud resides as $\Pi$, then we have $q \in \Pi$.


Now we denote an image sequence as: $\{(I_k, t_k)\}_{k=0}^{L-1}$,
where $L$ represents the length of the image sequence, $I_k$ represents the $k$-th, and $t_k$ is its associate time like age. The deformation occurring within an image over time is essentially a mapping from the original spatial positions at a starting point to the new spatial positions at the next time point, which can be represented as:
$\psi:\Omega\to\Omega$. 
On this basis, the identity mapping $Id = \psi_0$ can be defined as: $\psi_0(x)=x$, for all $x\in q$. In many applications, we desire the deformation field $\psi$ to possess the properties of smoothness and diffeomorphism.

The objective function of the deformable image registration is defined as:
\begin{equation}\label{eqn:registration_loss}
    \mathcal{J}(\psi;I_m,I_f)=\mathcal{S}(I_m(\psi(q_0)),I_f)+\mathcal{R}(\psi),
\end{equation}
where $q_0$ represents the initial voxel cloud without any deformation. The term $\mathcal{S}(\cdot,\cdot)$ denotes a similarity metric, used to measure the similarity between the moving image $I_m$ deformed by $\psi$ and the fixed image $I_f$. The term $\mathcal{R}(\cdot,\cdot)$ represents the regularization constraint applied to the deformation field. By generalizing image registration to the temporal regression, the objective function is updated as:
\begin{equation}\label{equ:regression_loss}
    \mathcal{J}(\psi; \{(I_k, t_k)\}_{k=0}^{L-1})=\sum_{k=1}^{L-1}(\mathcal{S}(I_0(\psi_k(q_0)),I_k)+\mathcal{R}(\psi_k)).
\end{equation}
In this way, we regress the image sequence and generate an image trajectory, where the generated images are as close as possible to the corresponding images in the original sequence, while imposing the smoothness constraints on the deformation fields.


\subsection{Formulation of Image Regression in Neural ODEs}
From a system perspective, neural ODEs represent vector fields as a continuous-time model of neural networks. It has been widely used as a general framework for modeling high-dimensional spatio-temporal chaotic systems using convolutional layers, demonstrating its ability to capture highly complex behaviors in space and time. Therefore, we consider the trajectory of the entire voxel cloud as the solution of the following first-order ordinary differential equation (ODE):
\begin{equation}\label{eq:5}
\frac{d q}{d t} =\mathbf{v}_\theta(q(t),t),  \quad 
s.t. \; q(0)=q_0=Id,
\end{equation}
where $\mathbf{v}_{\theta}(\cdot)$ is a parameterized network that describes the dynamics of voxel cloud deformation, $q_0$ represents the initial state of the voxel cloud at $t=0$, which corresponds to an identity map. The varying velocity field over time indicates non-stationary dynamics, which is fundamentally different from SGR with stationary dynamics.

The trajectory of $q$ is generated by integrating the above ODE under the initial condition $q_0$. Assuming the voxel cloud evolves from $t = 0$ to $t = t_k(k=1,2,...,L-1)$, the voxel cloud obtained at $t = t_k$ is given by the following equation:
\begin{equation}\label{eq:6}
\psi_k(q_0)=q(t_k)=q_0+\int_0^{t_k}\mathbf{v}_\theta(q(t),t)dt.   
\end{equation}
In particular, the computation of this flow field map is performed using numerical integration methods such as the Euler method~\cite{biswas2013discussion}. The time $t$ can be parameterized by the total number of steps and the corresponding step size adopted by the solver. Therefore, the task of finding the transformation $\psi$ becomes the search for the optimal parameter set $\theta$ that describes $\mathbf{v}$. The optimization problem becomes:
\begin{equation}\label{eq:7}
\theta=\underset{\theta\in\Theta}{\operatorname{arg}\min}
\sum_{k=1}^{L-1}
\left (\mathcal{S}\left(I_0(q_0+\int_0^{t_k}\mathbf{v}_\theta(q(t),t)dt), I_k\right) 
+ \mathcal{R}(\psi_k,\mathbf{v}_\theta)
\right),
\end{equation}
where $\Theta$ represents the entire parameter space. 
Since Neural ODEs typically require numerical solvers and take many steps to approximate flows, they will incur significant memory overhead if all gradients along the integration steps need to be stored during backpropagation. Hence, the Adjoint Sensitivity Method (ASM)~\cite{chen2018neural,pontryagin2018mathematical} has been implemented for optimizing Neural ODEs with constant memory gradient propagation, allowing our framework to interpolate any number of time steps between $t = 0$ and $t = s$ with a constant memory overhead.

\noindent
\textbf{Latent Space.}
Due to the complexity of high-dimensional data, solving Neural ODEs directly in the original space incurs significant computational costs. Therefore, we bring the above image regression formulation into a latent space, using a pair of pre-trained encoder-decoder networks to reduce the dimension of deformations, as shown in Fig.~\ref{fig:overview}. We apply diffeomorphic VoxelMorph~\cite{dalca2018unsupervised,dalca2019unsupervised} on a large 3D MRI dataset to estimate the deformations between image pairs and use these estimated deformations to guide the pre-training of the auto-decoder. At the training stage of our regression model, we fine-tune the decoder. Overall, the final framework of our NODER can be represented as:
\begin{equation}\label{eq:8}
\begin{aligned}
\frac{d y}{d t} =\mathbf{u}_\theta(y(t),t),  
s.t. \quad &y(0)= \text{Encoder}(q_0), \quad
y(t_k)=y_0+\int_0^{t_k}\mathbf{u}_\theta(y(t),t)dt., \\
&q(t_k)=\mathcal{K}(\text{Decoder}(y(t_k))),
\end{aligned}
\end{equation}
where Encoder$(\cdot)$ and Decoder$(\cdot)$ represent the encoder and decoder, respectively, following the design in~\cite{shu2018deforming}. $\mathcal{K}$ denotes a smoothing kernel used to smooth the deformation fields obtained after decoding. $\mathbf{u}_\theta$ is a parameterized network like $\mathbf{v}_\theta$ but in the latent space, which is used to estimate dynamics. In the dynamic network $\mathbf{u}_\theta$, we extract features from the latent space through continuous convolutional downsampling. These features are then flattened into one-dimensional vectors and added to the input time embedding, achieving fusion between the latent space features and time. Finally, we reconstruct the output of the fully connected layers, restoring the one-dimensional vector to the shape of the compressed three-dimensional deformation field. The overview of our model is presented in Fig.~\ref{fig:overview}.

\noindent
\textbf{Loss Functions.} We choose the normalized cross-correlation (NCC) as the loss function for the similarity term $\mathcal{S}$.
The regularization term $\mathcal{R}$ consists of two parts:
\begin{equation}\label{eq:11}
\mathcal{R}(\psi)=\lambda_{1}\mathcal{L}_{smt}+\lambda_{2}\mathcal{L}_{bdr} = \lambda_{1}\frac{1}{N}\sum\limits_{x\in q(s)}(\|\nabla \psi(\mathrm x)\|_2^2) + 
\lambda_{2} \frac{1}{N_{bdr}}\sum\limits_{d\in D} \sum\limits_{b\in B} \|\psi_{d,b}(\mathrm x)\|_2^2 ,
\end{equation}
where the first term $\mathcal{L}_{smt}$ constrains the smoothness of the spatial gradients within the deformed voxel cloud, and the second term $\mathcal{L}_{bdr}$ represents the $L_2$-norm constraint on the boundary of the deformation field.
Here, $N$, $N_{bdr}$, $D = (d_1, d_2, d_3)$, and $B=(top,bottom,front,behind,left,right)$ represent the total voxel number, the voxel number of six boundary planes,the three dimensions, and the six boundary planes of the deformation field, respectively. 

\section{Experiments}
We conducted experiments on two medical datasets, including a 3D MRI brain image dataset ADNI (Alzheimer’s Disease Neuroimaging Initiative)~\cite{mueller2005ways} and the cardiac dataset ACDC~\cite{bernard2018deep}. We compare our method with two advanced medical image regression baselines, FPSGR~\cite{ding2019fast} and SADM~\cite{yoon2023sadm}. Finally, we perform ablation experiments to demonstrate the effects of a series of smoothness constraints within the network.


\begin{figure}[t]
\centering
\includegraphics[width=1.0\textwidth]{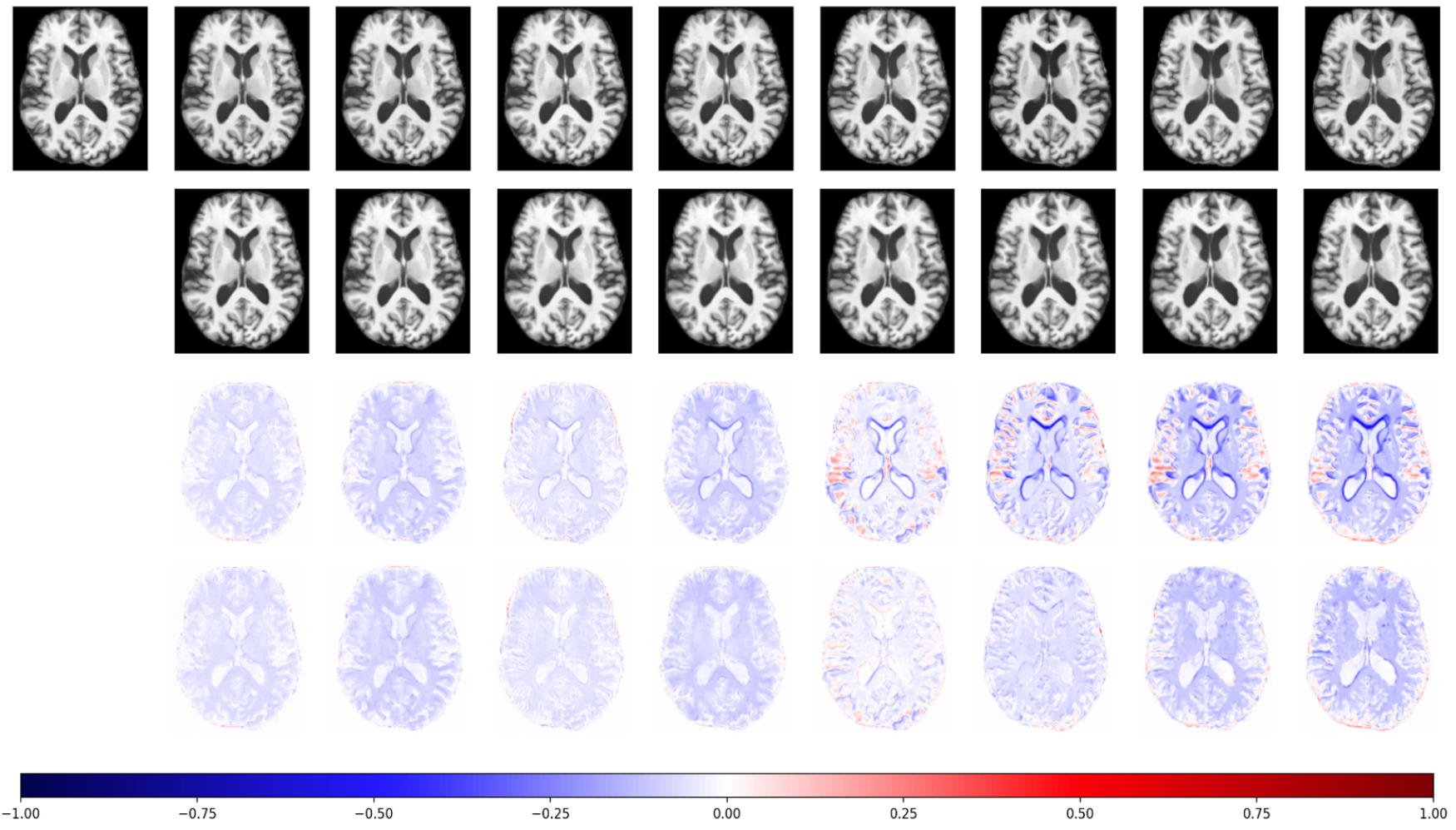}
\caption{Visualization of the regression results on the ADNI dataset. Top to bottom rows: 1) image sequence of a subject, 2) our prediction, 3) image difference of original sequence w.r.t. the first baseline image, and 4) image difference between generated and original corresponding images.} \label{ADNI-vs}
\end{figure}

\noindent
\textbf{Datasets.}
{\it (1) ADNI~\cite{mueller2005ways}.} The ADNI dataset consists of 3D brain MRI images collected from 2,334 subjects. Each subject has an image sequence of 1 to 16 time points, resulting in 10,387 MRI images. All images went through preprocessing steps including denoising, bias field correction, skull stripping, and affine registration to the SRI24 atlas. All brain images are standardized to a size of $144\times 176 \times 144$ with a spacing of $1mm\times 1mm \times 1mm$ and applied histogram equalization. The intensity of each image volume is normalized within $[0,1]$. We select 1,568 subjects that have more than two image scans as our dataset for experiments. For each subject, we randomly select 20\% time points for the test and the remaining images are used for training and validation. 
{\it (2) ACDC~\cite{bernard2018deep}.} The ACDC (Automatic Cardiac Diagnosis Challenge) dataset consists of cardiac MRI images from 100 training subjects and 50 testing subjects. We follow SADM~\cite{yoon2023sadm} and borrow its pre-processed and partitioned ACDC dataset. We take the image sequence from the ED (the End-Diastole of the cardiac cycle) to the ES (End-Systole) and resize it to 12 image frames and each frame has a size of $128\times 128\times 32$.

\begin{table}[t]
\centering
\tabcolsep=0.1cm
\caption{Quantitative comparison between baselines and our propose NODER} 
\label{tab:metrics-compare}
\begin{tabular}{c|c|ccc|c|cc}
\toprule
Dataset & Method      & NRMSE $\downarrow$ & SSIM $\uparrow$ & PSNR $\uparrow$ & \%Foldings $\downarrow$ & Inf. time $\downarrow$ & Memory $\downarrow$\\ \midrule
\multirow{3}{*}{ACDC} & FPSGR~\cite{ding2019fast}       &0.321        & 0.674    &22.852      & \textbf{5.9e-4}          & \textbf{1.58s}  & \textbf{5.6GB}          \\ 
& SADM~\cite{yoon2023sadm}        &0.287        &0.701      &24.996      & --            &5min       &38.5GB    \\ 
& NODER (ours) & \textbf{0.283}      & \textbf{0.712}      & \textbf{25.547}     & 2.3e-3          &15.63s   &10.3GB        \\ \midrule
\multirow{2}{*}{ADNI} & FPSGR~\cite{ding2019fast}       &0.184        & 0.755     & 26.876     & \textbf{3.2e-4}          & \textbf{1.72s}  & \textbf{8.2GB}         \\ 
& NODER (ours) & \textbf{0.159}        & \textbf{0.842}      & \textbf{28.673}     & 1.8e-3          &18.53s  &15.4GB         \\ 
\bottomrule
\end{tabular}
\end{table}

\begin{figure}[t]
\centering
\includegraphics[width=1.0\textwidth]{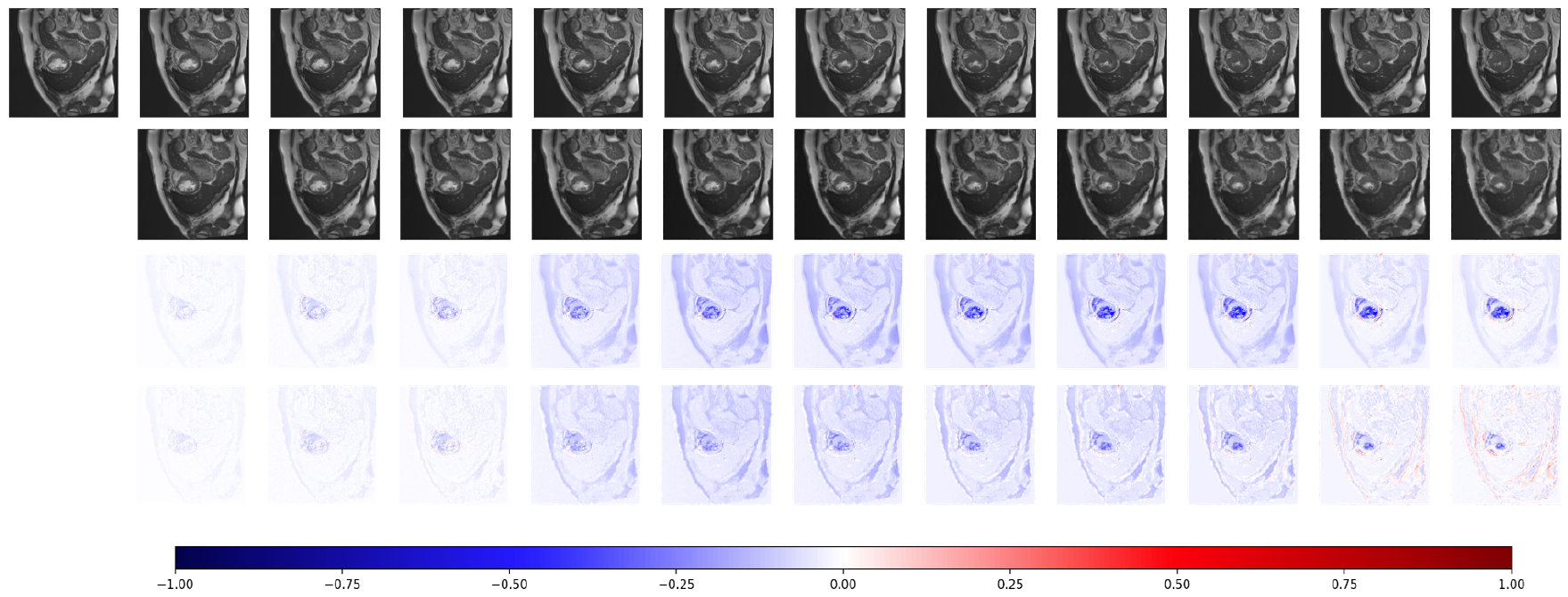}
\caption{Visualization of the regression results on the ACDC dataset. Top to bottom rows: 1) image sequence of a subject, 2) our prediction, 3) image difference of original sequence w.r.t. the first baseline image, and 4) image difference between generated and original corresponding images.} \label{ACDC-vs}
\end{figure}

\noindent
\textbf{Evaluation Metrics and Other Settings.}
To evaluate the quality of the generated images, we use three metrics, including the Normalized Root Mean Square Error (NRMSE), the Structural Similarity (SSIM), and the Peak Signal-to-Noise Ratio (PSNR). Also, we quantify the smoothness of a deformation field by calculating the percentage of its voxels with negative Jacobian determinants. Regarding the inference time, we implement our models and FPSGR on a single RTX 3090 GPU and report their memory cost and the average time of 5 forward inferences. Since SADM needs more GPU memory, we implement it on A100-PCIE-40GB GPU and then report its computational cost. For other inference costs, we load the model on a single RTX 3090 GPU and record the time and storage required for one forward inference, averaging multiple forward inferences to obtain the average costs.

For the ACDC dataset, we compare our method with both FPSGR and SADM, while on the ADNI dataset, we have only FPSGR as the baseline, since SADM cannot handle it even with an A100 GPU. Due to the lack of ground truth and other technique issues, we replace the registration networks of FPSGR with the diffeomorphic VoxelMorph~\cite{dalca2019unsupervised}, which is pre-trained on the ADNI dataset.

In the specific implementation of NODER, we choose the average smoothing kernel with a window size of 15 and a sliding stride of 1. The optimizer is Adam, with a learning rate set to 0.005. The construction of Neural ODE relies on the torchdiffeq toolkit~\cite{torchdiffeq}, where the solving method is set to RK4 (fourth-order Runge-Kutta method with a fixed step size). The relative error tolerance (rtol) is set to 1e-3, and the absolute error tolerance (atol) is set to 1e-5. The coefficients $\lambda_1$ and $\lambda_2$ for the loss functions $\mathcal{L}_{smt}$ and $\mathcal{L}_{bdr}$  are set to 0.05 and 0.0001, respectively. For image sequences from a single subject, we train our model for a total of 300 epochs.






\begin{figure}
\centering
\includegraphics[width=1.0\textwidth]{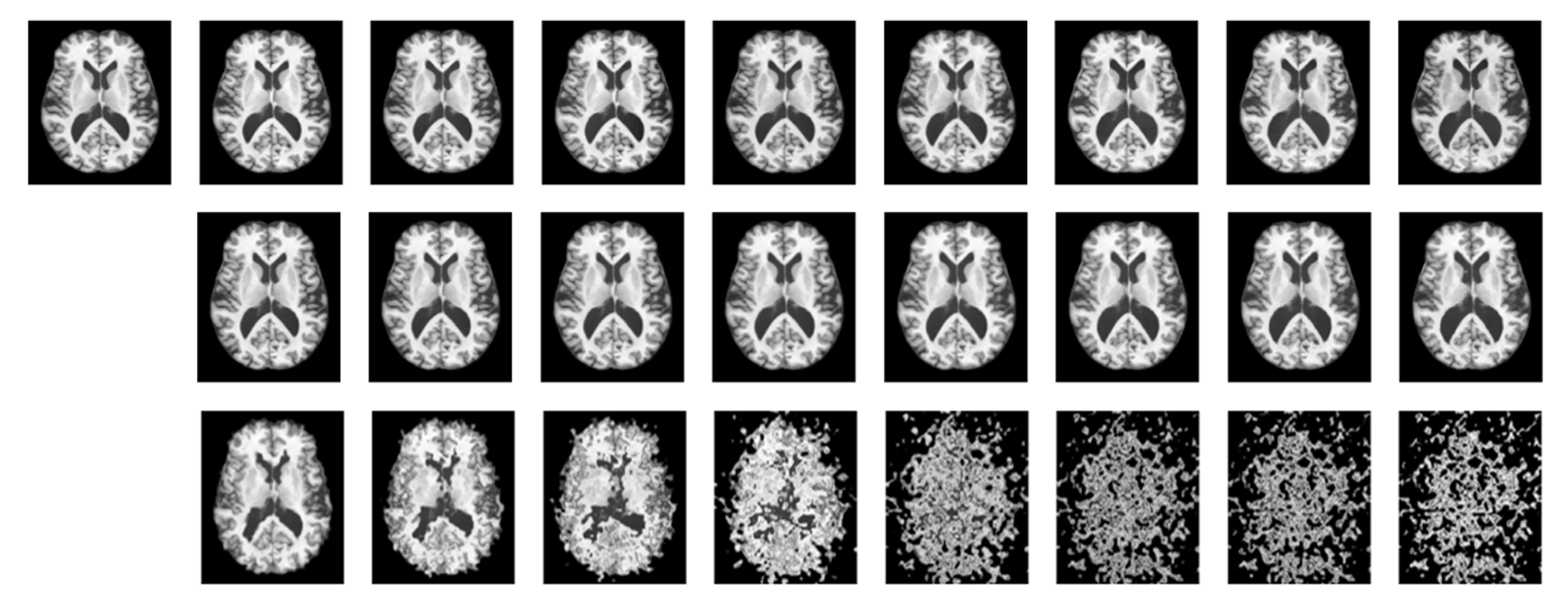}
\caption{Visualization of using (2nd row) and not using (3rd row) the smoothness constraints on regression the image sequence (1st row).} 
\label{ablation-smooth}
\end{figure}

\begin{wrapfigure}{r}{0.45\textwidth}
\centering
\includegraphics[width=0.43\textwidth]{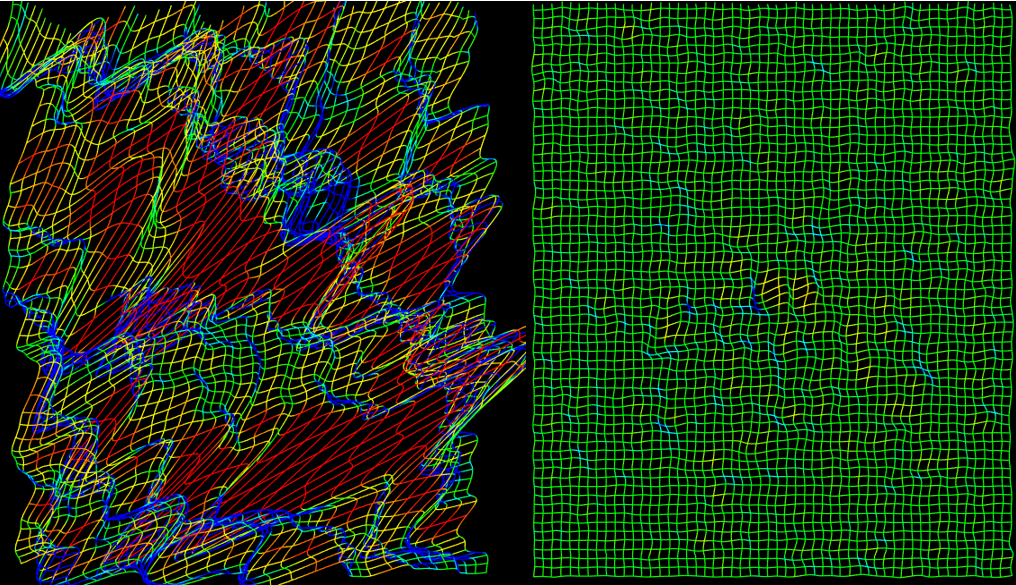}
\caption{Visualization of deformation fields without (left) and with the boundary condition (right).} 
\label{ablation-bdr}
\end{wrapfigure}

\noindent
\textbf{Experiment Results.} Table~\ref{tab:metrics-compare} reports the quantitative results of our method compared with FPSGR and SADM. Our NODER outperforms all methods in terms of the quality of the generated images. Our method needs more computational time and memory than FPSGR to generate images with higher quality, while both have the inference time within seconds and a memory cost of around 10GB. While SADM needs way more time and memory at the inference stage. 
The visualization of the regression results is shown in Fig.~\ref{ADNI-vs} and Fig.~\ref{ACDC-vs}. The difference images indicate our method can successfully capture the temporal dynamics in the brain and cardiac image sequences.

\noindent
\textbf{Ablation Study.} 
To validate the effect of the smoothness constraints in our proposed method, we conduct an ablation experiment, which is shown in fig.~\ref{ablation-smooth}.
By removing all smoothness constraints, it can be observed that the quality of the generated image sequence significantly deteriorates. Voxels move arbitrarily in three-dimensional space, leading to severe distortion of the original brain structure and numerous folding phenomena.
To verify the effect of the boundary condition $\mathcal{L}_{bdr}$ in $R(\psi)$, we visualize the deformation field, as shown in Fig.~\ref{ablation-bdr}. The introduction of boundary conditions leads to a powerful smoothness constraint in the background region of the deformation field, where severe deformations originally occurred. 

\section{Conclusion and Discussion}
In this work, we propose the NODER method, leveraging the powerful representation capability of neural networks to simulate the underlying dynamics of brain or cardiac deformation trajectories. Through solving ordinary differential equations, we achieve fitting regression on existing medical image time series, thus enabling the generation of desired images at any time point. Our method is based on the theoretical basis of deformable registration and resamples the first image of each subject through the deformation field to generate a new image. The loss terms we use in this paper explicitly impose diffeomorphic constraints, thus maintaining accurate anatomical information to some extent. Experimental results on large-scale 3D MRI datasets demonstrate that our method outperforms existing state-of-the-art methods, FPSGR and SADM, by predicting more accurate image volumes. In future work, we consider incorporating the learning-based methods to further reduce the inference time cost and make it more practical. Also, exploring ways to improve the smoothness of the deformation fields is another direction for our future work.

\subsubsection*{Acknowledgments}
\small{This work was supported by the National Natural Science Foundation of China (NSFC) 62203303 and Shanghai Jiao Tong University ``Jiao Da Star" Program for Interdisciplinary Medical and Engineering Research YG2022QN016 and YG2022QN028.}

\subsubsection*{Disclosure of Interests}
\small{The authors have no competing interests to declare that are relevant to the content of this article.}

\bibliographystyle{splncs04}
\bibliography{Paper-1899}

\end{document}